\documentclass[10pt,twocolumn,letterpaper]{article}

\usepackage{iccv}
\usepackage{times}
\usepackage{epsfig}
\usepackage{graphicx}
\usepackage{amsmath}
\usepackage{amssymb}
\usepackage{physics}
\usepackage{adjustbox}
\usepackage{subfig}
\usepackage{caption}
\usepackage{bm}
\usepackage{booktabs}
\usepackage{multirow}
\usepackage[dvipsnames]{xcolor}
\usepackage{diagbox}

\usepackage[breaklinks=true,bookmarks=false]{hyperref}

\iccvfinalcopy 

\def\iccvPaperID{31} 

\ificcvfinal\fi

\begin{document}

\newcommand{\projectName}{ZiCo-BC}
\title{\projectName: A \underline{B}ias \underline{C}orrected Zero-Shot NAS for Vision Tasks}

\author{Kartikeya Bhardwaj\thanks{Equal Contribution. $^\dag$Qualcomm AI Research is an initiative of Qualcomm Technologies,
Inc.} , Hsin-Pai Cheng$^*$, Sweta Priyadarshi$^*$, Zhuojin Li\\
Qualcomm AI Research$^\dag$ \\
San Diego, CA, USA 92121\\
{\tt\small \{kbhardwa, hsinpaic, swetpriy, zhuoli\}@qti.qualcomm.com}
}

\maketitle
\ificcvfinal\thispagestyle{empty}\fi

\begin{abstract}\vspace{-4mm}
Zero-Shot Neural Architecture Search (NAS) approaches propose novel training-free metrics called zero-shot proxies to substantially reduce the search time compared to the traditional training-based NAS. Despite the success on image classification, the effectiveness of zero-shot proxies is rarely evaluated on complex vision tasks such as semantic segmentation and object detection. Moreover, existing zero-shot proxies are shown to be biased towards certain model characteristics which restricts their broad applicability. In this paper, we empirically study the bias of state-of-the-art (SOTA) zero-shot proxy ZiCo across multiple vision tasks and observe that ZiCo is biased towards thinner and deeper networks, leading to sub-optimal architectures. To solve the problem, we propose a novel bias correction on ZiCo, called \projectName. Our extensive experiments across various vision tasks (image classification, object detection and semantic segmentation) show that our approach can successfully search for architectures with higher accuracy and significantly lower latency on Samsung Galaxy S10 devices.
\end{abstract}

\vspace{-4mm}
\section{Introduction}\vspace{-1mm}

Neural Architecture Search (NAS) algorithms have been widely used to automatically design highly accurate and efficient model architectures within a given search space. However, such techniques can be very computationally expensive as they require a lot of training resources. To address this limitation, Zero-Shot (Training-Free) NAS ~\cite{zico, zennas, deepmad, nnmass, maedet} has emerged recently, which relies on certain properties of neural network architectures to rank various models during the search without any actual training. As a result, these methods significantly accelerate the model searching process, enabling the identification of high-performing models more efficiently~\cite{zico, zennas, rans, nnmass, maedet}. 
\begin{figure}[t]
  \centering
   \includegraphics[width=1.0\linewidth]{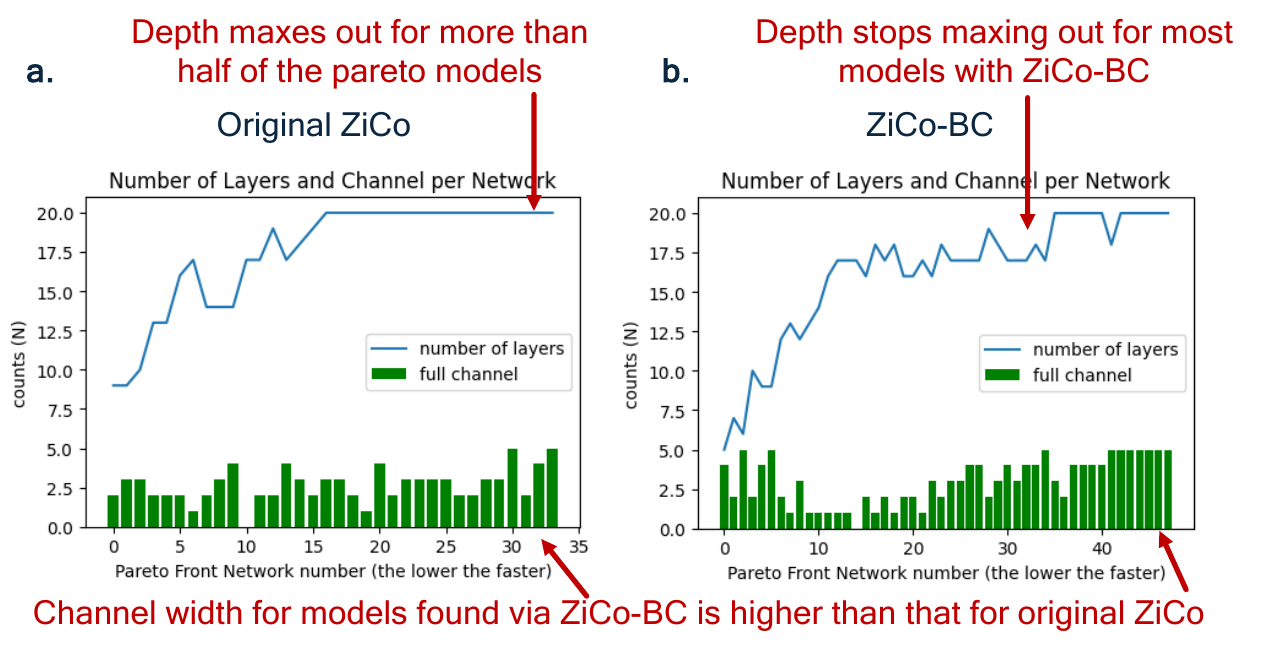}\vspace{-3mm}
   \caption{Overview: Zero-Shot NAS on ImageNet for EfficientNet type networks. (a)~ZiCo found architectures saturate with depth and have lower channel widths, thus showing that ZiCo is biased towards thinner and deeper networks. (b)~Bias-Corrected ZiCo-BC metric significantly reduces the depth-width bias and produces better models.}
   \label{fig:zicoV1V2}
\end{figure}

In this paper, we investigate two significant aspects of zero-shot NAS research. Firstly, despite the abundance of results in tasks such as image classification and various NAS-Benches \cite{nasbench101, nasbench, nasbench201, nasbench301}, several existing training-free metrics lack adequate validation on complex vision tasks, including semantic segmentation or object detection~\cite{zico, zennas, unif, xiang2023zero}. Secondly, training-free metrics can be biased towards specific characteristics in neural architectures~\cite{naszero}. For instance, existing zero-shot NAS proxies can exhibit bias towards various factors, such as cell sizes, skip connections, convolutions, number of parameters, etc.~\cite{naszero}. 

To address the limitations mentioned above, we explore the following \textbf{key questions} centered around a recently introduced training-free metric known as ZiCo (Zero-Shot metric based on Inverse Coefficient of Variation on gradients)~\cite{zico} which has demonstrated state-of-the-art performance across various NAS-Benches and ImageNet task:
\begin{enumerate}\vspace{-1mm}
    \item Can ZiCo effectively perform \textit{direct search} in complex vision tasks, such as semantic segmentation or object detection without relying on initial ImageNet search? \vspace{-2mm}
    \item Are there any biases present in ZiCo? If yes, how can we correct these biases?
\end{enumerate}

Our study demonstrates that ZiCo yields exceptional results when applied to challenging semantic segmentation and object detection tasks, especially for \textit{macro-architecture search}. However, when conducting \textit{micro-architecture search} with a fixed backbone ~\cite{donna}), we observe a bias towards thinner (i.e., lower channel width) and deeper networks in ZiCo. Fig.~\ref{fig:zicoV1V2} demonstrates zero-shot NAS results for ImageNet on a broader search space than that considered in~\cite{zico}. As evident, the original ZiCo score tends to favor architectures with maximum depth and lower widths, leading to a bias towards thinner and deeper networks. This bias can hinder the effectiveness of zero-shot NAS methods across various applications, as it may lead to sub-optimal neural networks with lower accuracy. Therfore, there is a need for bias correction methods that can significantly improve the performance of zero-shot NAS.

In summary, we make the following \textbf{key contributions}: (1)~We demonstrate that gradient-based zero-shot proxies like ZiCo are capable of performing \textit{direct} macro-architecture searches on complex vision tasks such as semantic segmentation and object detection. (2)~We propose a new bias correction method for ZiCo, called ZiCo-BC, that significantly enhances the metric's performance and identifies effective models in micro-architecture searches.
 (3)~Finally, we also provide general guidelines on how to scale up this bias correction for ZiCo prior to training individual models, along with an assessment of its current limitations.

\section{Zero-Shot NAS for Complex Vision Tasks}
\paragraph{Preliminaries.} \vspace{-2mm}
ZiCo~\cite{zico} is a zero-shot NAS metric that leverages the inverse coefficient of variation on gradients. This score is used to rank neural network models based on their convergence rate and generalization capacity. Specifically, ZiCo is computed as follows~\cite{zico}: \vspace{-2mm}
\begin{equation} \vspace{-1mm}
    \text{ZiCo} = \sum_{l=1}^{D} \text{log} \left( \sum_{\theta_l} \frac{\mathbb{E}[\grad_{\theta_l}]}{\sqrt{\text{Var}(\grad_{\theta_l})}} \right), 
    \label{eq::zicoV1}
\end{equation}
where, $D$ is the total number of layers in the network, $\theta_l$ represents each parameter in layer $l\in \{1,2,3,\ldots,D\}$, and $\grad_{\theta_l}$ is the gradient of the loss w.r.t. each parameter $\theta_l$. The expected value and standard deviation is computed across multiple batches of input data at initialization. That is, no parameters are updated across batches, only forward and backward passes are used to compute gradient statistics. It was theoretically shown in~\cite{zico} that these gradient statistics are linked to training convergence and generalization.

In this paper, we will discuss two kinds of zero-shot NAS paradigms using ZiCo: \textbf{(1)~Macro-Architecture Search} where we search over multiple types of backbones and heads; we call this macro-architecture search since it significantly impacts the topology of neural architectures, and \textbf{(2)~Micro-Architecture Search} where the backbone type is fixed and a same type of block repeats throughout the network; here, we search over channel counts, number of block repeats, kernel sizes, type of convolution (regular, depthwise, group), expansion ratios, etc.~\cite{donna}.

\paragraph{Does ZiCo work on complex vision tasks like semantic segmentation? A Direct Macro-Architecture Search.} Despite extensive theoretical contributions and empirical validation across several NAS-Benches and ImageNet, the effectiveness of ZiCo was not evaluated for \textit{direct search} over downstream computer vision tasks, i.e., without any prior ImageNet search. Hence, in this section, we exploit ZiCo to directly search for hardware-efficient Semantic Segmentation networks in a wide search space containing multiple types of backbones and segmentation heads.

We construct a complex search space using backbone and head from HRNet~\cite{hrnet} architecture as well as using backbones and heads from a recent \textit{manually-designed} hardware-efficient semantic segmentation network called FFNet~\cite{ffnet}. HRNet~\cite{hrnet} and FFNet~\cite{ffnet} are highly different architectures. We also searched for HRNet-Head~\cite{hrnet} or the Up-B-Head from FFNet~\cite{ffnet} for head search. Finally, we introduced individual options for each backbone (e.g., depth, width, etc.), leading to a large search space. 

Our objective is to exploit ZiCo to automatically design a significantly better network than the manual FFNet which was designed for mobile-scale AI accelerators. To this end, we consider the Cityscapes segmentation task and conduct NSGA-2 evolutionary search~\cite{nsga2} over the above search space with hardware latency in the loop on the Samsung Galaxy S10 mobile platform. For ZiCo computation, we used the same loss as the one used to train FFNet~\cite{ffnet}.
\begin{table}[]
\caption{Direct \textbf{Macro-Architecture Search} via ZiCo on Cityscapes Semantic Segmentation}\vspace{-3mm}
\centering
\scalebox{0.77}{
\begin{tabular}{|l||c|c||c|c|}
\hline
Model & \textit{\#}Params & \textit{\#}MACs & Latency (ms) & mIoU\\ \hline
HRNet~\cite{hrnet} & \textbf{3.94M}  & \textbf{77.89G}  & 28.80 ($1\times$) & 77.0\% \\ \hline
FFNet~\cite{ffnet} &  27.49M & 96.37G  & \textbf{8.35} ({\color{ForestGreen}{\bm{$3.4\times$}\bm{$\downarrow$}}})  &  79.7\% ({\color{ForestGreen}{$+2.7\%$}})  \\ \hline \hline
\textbf{ZiCo model} &  31.92M  &  96.14G & \textbf{8.48} ({\color{ForestGreen}{\bm{$3.4\times$}\bm{$\downarrow$}}})  &  \textbf{80.7\%} ({\color{ForestGreen}{\bm{$+3.7\%$}}}) \\ \hline
\end{tabular}
}
\label{tab:zicoHRNet}
\end{table}

Table~\ref{tab:zicoHRNet} demonstrates the search results. As evident, even though the HRNet architecture has the least number of parameters and MACs, the HRNet backbone is not friendly to constrained mobile-scale hardware and shows about $3.4\times$ higher latency compared to the manual FFNet and our automatically found ZiCo-based model, both achieving much higher accuracy. Clearly, the ZiCo-based model significantly outperforms the manual FFNet by $1\%$ higher mIoU on Cityscapes segmentation with a similar latency.

\vspace{-1mm}
\section{Proposed Bias Correction}\vspace{-1mm}
As mentioned earlier, we observed a bias in ZiCo towards thinner and deeper networks for micro-architecture search, i.e., when similar blocks repeat themselves throughout the fixed backbone. More precisely, in equation \eqref{eq::zicoV1}, the metric sums over the number of layers in the network. Consequently, the score grows linearly in number of layers, whereas the gradient statistics grow logarithmically. For networks with repeating blocks, this can lead to deeper models achieving higher ZiCo scores even if they have significantly lower width. However, thinner and deeper networks may not always achieve optimal accuracy. As shown in Bhardwaj et al.~\cite{rans}, width plays a fundamental role in model's expressive power, and due to bias towards thinner and deeper networks, zero-shot metrics can become less effective at identifying optimal architectures. Therefore, due to the bias, ZiCo can favor deeper and thinner models over potentially more optimal ones during the evolutionary search. In the rest of this paper, we will discuss how to correct this depth-width bias in ZiCo.

\paragraph{Bias Correction for Micro-Architecture Search.}
To rectify the bias in ZiCo or other training-free NAS metrics that may exhibit a preference for thinner and deeper networks, we introduce a bias correction term. This term can be applied to modify the original metric definition. The proposed bias correction equation takes into account the ~\textit{feature map resolution} and ~\textit{channel width} of the network at different layers. For ZiCo, the equation is as follows:\vspace{-2mm}
\begin{equation}
{\small
\begin{aligned}
    \text{ZiCo-BC} &= \sum_{l=1}^{D} \text{log} \left( \left[\frac{H_l W_l}{\sqrt{C_l}}\right]^{-\beta} \sum_{\theta_l} \frac{\mathbb{E}[\grad_{\theta_l}]}{\sqrt{\text{Var}(\grad_{\theta_l})}} \right) \\
    &= \sum_{l=1}^{D} \text{log} \left( \sum_{\theta_l} \frac{\mathbb{E}[\grad_{\theta_l}]}{\sqrt{\text{Var}(\grad_{\theta_l})}} \right) 
    - \beta \sum_{l=1}^{D} \text{log}\left( \frac{H_l W_l}{\sqrt{C_l}} \right)\\
    &= \text{ZiCo}- \beta \sum_{l=1}^{D} \text{log}\left( \frac{H_l W_l}{\sqrt{C_l}} \right)
    \label{eq::zicoV2}
\end{aligned}
}
\end{equation}\vspace{-4mm}

\vspace{-1mm}
\noindent
Here, $H_l, W_l, C_l$ are height, width of the feature map, and number of channels in layer $l$, respectively. Hyperparameter $\beta$ controls the amount of depth-width penalty applied to the score. Setting $\beta=0$ automatically yields the original ZiCo score. Clearly, if the model becomes deeper or if it has fewer channels, the penalty increases, thus, discouraging thinner and deeper models during the evolutionary search. Of note, other bias correction methods may be possible. We comment on this briefly in Section 5. 

\section{Experiments} \vspace{-2mm}

We first conduct a NATS-Bench-SSS study on CIFAR-10, CIFAR-100, and ImageNet-16-120 datasets \cite{imagenet16-120} to evaluate the correlations of the proposed ZiCo-BC score with accuracy and compare them to the original ZiCo score. We then evaluate the proposed bias correction for three computer vision applications: (1)~ImageNet Image Classification, (2)~MS COCO Object Detection, and (3)~Cityscapes Semantic Segmentation. We use ResNet-based search space for semantic segmentation, and EfficientNet-based search space for ImageNet image classification as well as object detection. Next, we present more details on the micro-architecture search space and evolutionary search settings as well as performance of ZiCo-BC for each task. 

\vspace{-1mm}
\subsection{NAS-Bench Correlations }\vspace{-1mm}

Firstly, we evaluate the proposed bias correction on NAS benchmark NATS-Bench \cite{nasbench}. Specifically, we focus on the 32768 neural architectures with varying channel sizes from ``size search space'' (NATS-Bench-SSS), which resembles our micro-architecture search setting. 
Following the experimental setup in ZiCo, we compute the correlation coefficients (i.e., Kendall’s $\tau$ and Spearman’s $\rho$) between the zero-shot proxy and the test accuracy. As evident from Table \ref{tab:zicoBC_nasbench}, the bias correction improves the correlation score of ZiCo across all three datasets, indicating that the ZiCo-BC score can be a more representative proxy of test accuracy for ranking candidates during a micro-architecture search. 

\begin{table}[t]
\caption{Correlation Coefficients on NATS-Bench-SSS}\vspace{-2mm}
\centering
\scalebox{0.8}{
\begin{tabular}{|l||c|c||c|c||c|c|}
\hline
Dataset & \multicolumn{2}{|c||}{Cifar-10}  &  \multicolumn{2}{|c||}{Cifar-100} &  \multicolumn{2}{|c|}{Img16-120} \\ \hline\hline
\diagbox[width=5.5em]{Proxy}{Corr.} & KT  &  SPR & KT & SPR & KT & SPR    \\ \hline
ZiCo & 0.72  & 0.91 & 0.56  & 0.76 & 0.73 & 0.90  \\ \hline \hline
\textbf{ZiCo-BC} &  \textbf{0.78}  & \textbf{0.94}  &  \textbf{0.60} &  \textbf{0.79} & \textbf{0.79} & \textbf{0.94} \\ \hline
\end{tabular}
}
\label{tab:zicoBC_nasbench}
\end{table}

\subsection{Classification and Object Detection}
\vspace{-1mm}
We conduct classification and object detection tasks on EfficientNet and EfficientDet, respectively~\cite{efficientnet,efficientdet}. As these two networks share very similar backbones, we build the search space defined in previous studies~\cite{donna, efficientnet}. Specifically, the search space includes: (1) kernel size ($3\times3$ or $5\times5$), (2) channel size, (3) the number of operation repeats per block, and (4) regular convolution or group convolution (with group size of 32).
It is worth noting that one significant difference in our search space, as compared to existing works, is the omission of the squeeze-and-excite operation as it is not very hardware-friendly.
When incorporating ZiCo-BC into the search process, we utilize the widely employed cross-entropy loss~\cite{efficientnet} for classification and the focal loss~\cite{efficientdet} for object detection, respectively.

\vspace{-2.5mm}
\paragraph{Classification.}
EfficientNet~\cite{efficientnet} has proven to be a powerful architecture, achieving remarkable results in various computer vision tasks. To showcase the effectiveness of our proposed method, we employ ZiCo-BC to conduct a search for EfficientNet style models on the challenging ImageNet-1k dataset. 
By applying ZiCo-BC to this architecture, we aim to further enhance its performance and explore architectures that strike a balance between model depth and width. 
The model discovered by ZiCo-BC achieves an impressive 11\% reduction in latency without sacrificing accuracy, as demonstrated in Table \ref{tab:zicoBC_effnet}. In contrast, the original ZiCo score loses about $0.9\%$ accuracy for similar latency.

\vspace{-2.5mm}
\paragraph{Object Detection.}
EfficientDet \cite{efficientdet} is a family of architectures renowned for high accuracy and efficiency in object detection, accommodating various resource constraints.
The EfficientDet-D0 architecture comprises three components: (1) an EfficientNet backbone network, (2) a weighted bi-directional feature pyramid network (BiFPN), and (3) a class and box network for predicting object class and bounding box information. Notably, the backbone of EfficientDet-D0 contributes 78\% of the FLOPs and 92\% of the parameters in the entire architecture. Hence, our primary focus lies in searching for a backbone network that enhances the latency of the architecture without sacrificing accuracy.
Table \ref{tab:zicoBC_effnet} displays the results on MS COCO 2017 \cite{mscoco} after training for 300 epochs. Our searched architecture achieves a remarkable 29\% latency reduction while maintaining even better accuracy compared to EfficientDet-D0.

\begin{table}[t]
\caption{Direct \textbf{Micro-Architecture Search} via ZiCo and ZiCo-BC on EfficientNet/Det search space}\vspace{-3mm}
\centering
\scalebox{0.8}{
\begin{tabular}{|l||c||c|c|}
\hline
Model & Approach & Latency (ms) & Accuracy/mAP \\ \hline
\multirow{3}{*}{EfficientNet}    & Scaling   & 0.90  & 77.7\%  \\ \cline{2-4}
                        & ZiCo     &  0.82({\color{ForestGreen}{$-8\%$}})  &  76.8\% ({\color{red}{$-0.9\%$}})  \\ \cline{2-4}
                        &  \textbf{ZiCo-BC}     & \textbf{0.80 ({\color{ForestGreen}{\bm{$-11\%$}}})}  & \textbf{77.7\%} ({\color{ForestGreen}{\bm{$0\%$}}})  \\ \hline\hline

\multirow{2}{*}{EfficientDet}    & Scaling   &  2.792 & 33.6   \\ \cline{2-4}

                        &  \textbf{ZiCo-BC}     &  \textbf{1.974 ({\color{ForestGreen}{\bm{$-29\%$}}})}  &  33.8 ({\color{ForestGreen}{\bm{$+0.2\%$}}})  \\ \hline
\end{tabular}
}
\label{tab:zicoBC_effnet}
\end{table}

\vspace{-1mm}
\subsection{Semantic Segmentation}\vspace{-1mm}
In this section, we evaluate the bias correction ability of the proposed ZiCo-BC score in the context of micro-architecture search on Cityscapes dataset. Unlike Section 2, where we conducted a macro-architecture search across HRNet~\cite{hrnet} and FFNet~\cite{ffnet}, here we specifically test ZiCo-BC on the FFNet backbone in conjunction with the FFNet-Head. The FFNet backbone is based on the ResNet architecture and consists of four stages. 
Our micro-architecture search space consists of (1)~number of residual blocks in each stage, (2)~number of output channels for each stage; each residual block in the stage has the same number of channels, and (3)~type of convolution, i.e., Group Convolution with a group size of 32/64/128 channels whichever is larger, or a Regular Convolution. All kernel sizes are fixed to $3\times3$. 
To search over a large space, we significantly vary the width and depth of the candidate networks around the baseline FFNet~\cite{ffnet} configuration. Overall, this search space consists of more than 44M unique architectures. 

Table~\ref{tab:zicoBCFFNet} shows that ZiCo-BC finds a model with similar mIoU as FFNet~\cite{ffnet} but achieves 11\% lower latency on the mobile platform. In contrast, networks found via the original ZiCo metric lose nearly 1\% mIoU with about 16\% lower latency. The ZiCo-BC model has 74 residual blocks with higher channel widths, thus, correcting the bias towards deeper and thinner networks. Improving the latency of FFNet~\cite{ffnet} by 11\% (with similar mIoU) is highly non-trivial as it is already designed for mobile devices.
\vspace{-1mm}

\section{General Guidelines and Limitations}
\vspace{-1mm}
\paragraph{General Guidelines.}
One crucial aspect of the proposed bias correction is determining the value of the main hyperparameter $\beta$, which influences the penalty on depth and width. 
An appropriate $\beta$ value can be obtained by analyzing the architecture of Pareto models found during evolutionary search. 
For instance, in Fig.~\ref{fig:zicoV1V2}(a), most models exhibit maximum depth with low width, indicating the presence of bias. To address this, we gradually increase $\beta$ to encourage more diverse architectures with intermediate depth. For classification and object detection tasks, we used $\beta$ = 1. On the other hand, we used $\beta$ = 2 for semantic segmentation zero-shot micro-architecture search.

\begin{table}[]
\caption{Direct \textbf{Micro-Architecture Search} via ZiCo and ZiCo-BC on Cityscapes Semantic Segmentation}\vspace{-3mm}
\centering
\scalebox{0.78}{
\begin{tabular}{|l||c|c||c|c|}
\hline
Model & \textit{\#}Params & \textit{\#}MACs & Latency (ms) & mIoU\\ \hline
FFNet &  27.49M & 96.37G  & 8.35 &  79.70\%  \\ \hline
ZiCo &  \textbf{21.80M} & \textbf{75.89G}  & \textbf{7.02} ({\color{ForestGreen}{\bm{$-16\%$}}})  &  78.62\% ({\color{red}{$-1.08\%$}})  \\ \hline \hline
\textbf{ZiCo-BC} &  23.28M  &  79.85G & 7.44 ({\color{ForestGreen}{$-11\%$}})  &  \textbf{79.71\%} ({\color{ForestGreen}{\bm{$+0.01\%$}}}) \\ \hline
\end{tabular}
}
\label{tab:zicoBCFFNet}
\vspace{-1mm}
\end{table}

\vspace{-2mm}
\paragraph{Limitations.}
Two limitations of our current bias correction are identified. Firstly, the bias correction applies solely to micro-architecture search with repeated blocks. For macro-architecture search, gradient statistics for different backbones (topologies) can result in a similar score even if they have highly different number of layers\footnote{We observed this between FFNet and HRNet candidates: For networks with similar trainability, HRNet-based models with nearly half the layers often achieve comparable ZiCo to deeper FFNet-based models.}. Therefore, a common penalty to backbones with different depths would treat the shallower backbone unfairly. Hence, further research is needed to come up with a universal bias correction (if needed) for macro-architecture search. Secondly, the current bias correction assumes a fixed input size for candidate models, disregarding the potential gain in accuracy for various vision tasks by increasing the image size. Hence, future bias correction methods that maintain the overall score with increasing input size are of interest.

\section{Conclusion}\vspace{-1mm}
In this paper, we explore the effectiveness of zero-shot NAS on complex vision tasks beyond traditional image classification. Firstly, we validate an existing proxy, called ZiCo for \textit{macro-architecture search} in semantic segmentation. The ZiCo-based network achieves a remarkable $3.4\times$ speed up over HRNet through automatic search, with 1\% higher mIoU compared to a manually designed model of similar latency. Next, we identify biases in ZiCo for \textit{micro-architecture search} and propose ZiCo-BC, a novel bias correction method for depth-width biases in zero-shot metrics. 
Finally, we demonstrate that our bias correction enables ZiCo-BC to consistently achieve $11$-$30\%$ lower latency and $0.2$-$1.1\%$ higher accuracy compared to the models found via the original ZiCo for micro-architecture search on image classification, object detection, and segmentation. 

{\small
\bibliographystyle{ieee_fullname}
\bibliography{egbib}
}

\end{document}